\documentclass[runningheads]{llncs}
\usepackage[T1]{fontenc}
\usepackage{graphicx,verbatim}
\usepackage{hyperref}
\usepackage{color}

\usepackage{amsmath}
\usepackage{amssymb}

\usepackage{makecell}
\usepackage{multirow}
\usepackage{pifont}
\usepackage{colortbl}
\usepackage[table]{xcolor}

\usepackage{marvosym}

\begin{document}
\title{LEAF: Latent Diffusion with Efficient Encoder Distillation for Aligned Features in Medical Image Segmentation}
\titlerunning{LEAF: Latent Diffusion with Efficient Distillation for Aligned Features}

\author{Qilin Huang\inst{1} \and
Tianyu Lin\inst{2} \and
Zhiguang Chen\inst{1}\and
Fudan Zheng\inst{1}$^{\textrm{\Letter}}$}

%
\authorrunning{Q. Huang et al.}
%
\institute{School of Computer Science and Engineering, Sun Yat-sen University, China\\
\email{zhengfd5@mail.sysu.edu.cn}\\\and
Department of Biomedical Engineering, Johns Hopkins University, United States}
    
\maketitle              
\begin{abstract}
Leveraging the powerful capabilities of diffusion models has yielded quite effective results in medical image segmentation tasks. However, existing methods typically transfer the original training process directly without specific adjustments for segmentation tasks. Furthermore, the commonly used pre-trained diffusion models still have deficiencies in feature extraction. Based on these considerations, we propose LEAF, a medical image segmentation model grounded in latent diffusion models. During the fine-tuning process, we replace the original noise prediction pattern with a direct prediction of the segmentation map, thereby reducing the variance of segmentation results. We also employ a feature distillation method to align the hidden states of the convolutional layers with the features from a transformer-based vision encoder. Experimental results demonstrate that our method enhances the performance of the original diffusion model across multiple segmentation datasets for different disease types. Notably, our approach does not alter the model architecture, nor does it increase the number of parameters or computation during the inference phase, making it highly efficient. Project page: \url{https://leafseg.github.io/leaf/}

\end{abstract}
\section{Introduction}

The diffusion model\cite{Sohl-DicksteinW15} has achieved successful results in multiple image generation tasks, demonstrating itself as a scalable approach to generate high-dimensional visual data. Due to its powerful capabilities, recent studies have begun exploring its potential for application in other vision tasks. For example, DMP\cite{0001T0024a} adapts a text-to-image diffusion model to obtain faithful estimations on several dense prediction tasks. Marigold\cite{KeOHMDS24} directly fine-tunes Stable Diffusion\cite{RombachBLEO22} for image-conditioned depth generation, achieving state-of-the-art (SOTA) performance on multiple depth estimation datasets while enabling zero-shot transfer to unseen data.

Given the powerful capabilities of diffusion models, numerous studies have explored their application to medical image segmentation\cite{MedSegDiff,MedSegDiffV2,DiffUNet}, demonstrating their remarkable potential and sparking growing research interest in the community. However, these methods typically directly adopt the original diffusion model training process and often incorporate overly complex modules to enhance feature representation. While enhancing performance, these designs bring about computational inefficiencies and obscure the fundamental differences between segmentation and generation tasks. In contrast, SDSeg\cite{LinCYYZ24} employs a latent diffusion model and improves inference speed and accuracy by utilizing a single-step reverse process. Nevertheless, this approach still fails to address the inherent divergence between segmentation objectives (e.g., pixel-wise classification) and generative modeling principles (e.g., noise prediction). Several prior works have investigated alternative parameterization methods\cite{SalimansH22,BennyW22} to generate detailed and realistic natural images. However, these approaches either rely on multi-step progressive generation or yield comparable evaluation metrics, thereby offering limited insights for this task.

Meanwhile, the widely used pre-trained diffusion models are based on a convolutional U-Net architecture. Many studies have pointed out that Transformer architectures\cite{VaswaniSPUJGKP17} can effectively enhance feature extraction, although they also increase both the computational cost and the number of parameters. In addition, some research\cite{LiPDBP23} indicates that while estimating pixel-level geometric attributes from a single image requires a comprehensive understanding of the scene, merely predicting results in the input space is insufficient for learning a robust representation. Consequently, achieving a good trade-off between speed and performance remains a significant challenge in current work. Distillation methods offer a promising solution. Notably, the recent REPA approach\cite{yu2024repa} accelerates model convergence by aligning the features of two Transformers\cite{VaswaniSPUJGKP17}, leading to improved generation performance.

Motivated by these concerns, we propose \textbf{LEAF} (\textbf{L}atent Diffusion with \textbf{E}fficient Encoder Distillation for \textbf{A}ligned \textbf{F}eatures). We analyze the diffusion formulation and discover that using noise prediction in segmentation tasks might not be optimal, as it tends to amplify prediction errors. Therefore, we replace this approach with one that directly predicts the sample. Moreover, we implement a novel and simple distillation method to enhance the feature representation of convolutional networks, allowing the model to align its features with those obtained from powerful Transformer-based models. Such alignment enhances segmentation performance without incurring any additional computational overhead or parameter increase during inference.

In conclusion, our main contributions are as follows:
\begin{itemize}
    \item We replace the high-variance $\epsilon$-prediction, originally used in diffusion models for generation tasks, with the low-variance $x_{0}$-prediction that is better suited for segmentation tasks, and provide the corresponding mathematical explanation.
    \item We design an efficient feature alignment method that enriches the representation of U-Net by distilling a powerful visual encoder, thereby improving segmentation performance on multiple medical imaging datasets across various diseases.
    \item Our method allows the alignment module to be removed during inference, incurring no additional computational or memory overhead. Moreover, this plug-and-play approach does not alter the internal structure of the model and can be easily transferred to other diffusion-based models.
\end{itemize}

\begin{figure}[htb]
    \centering
    \includegraphics[scale=0.36]{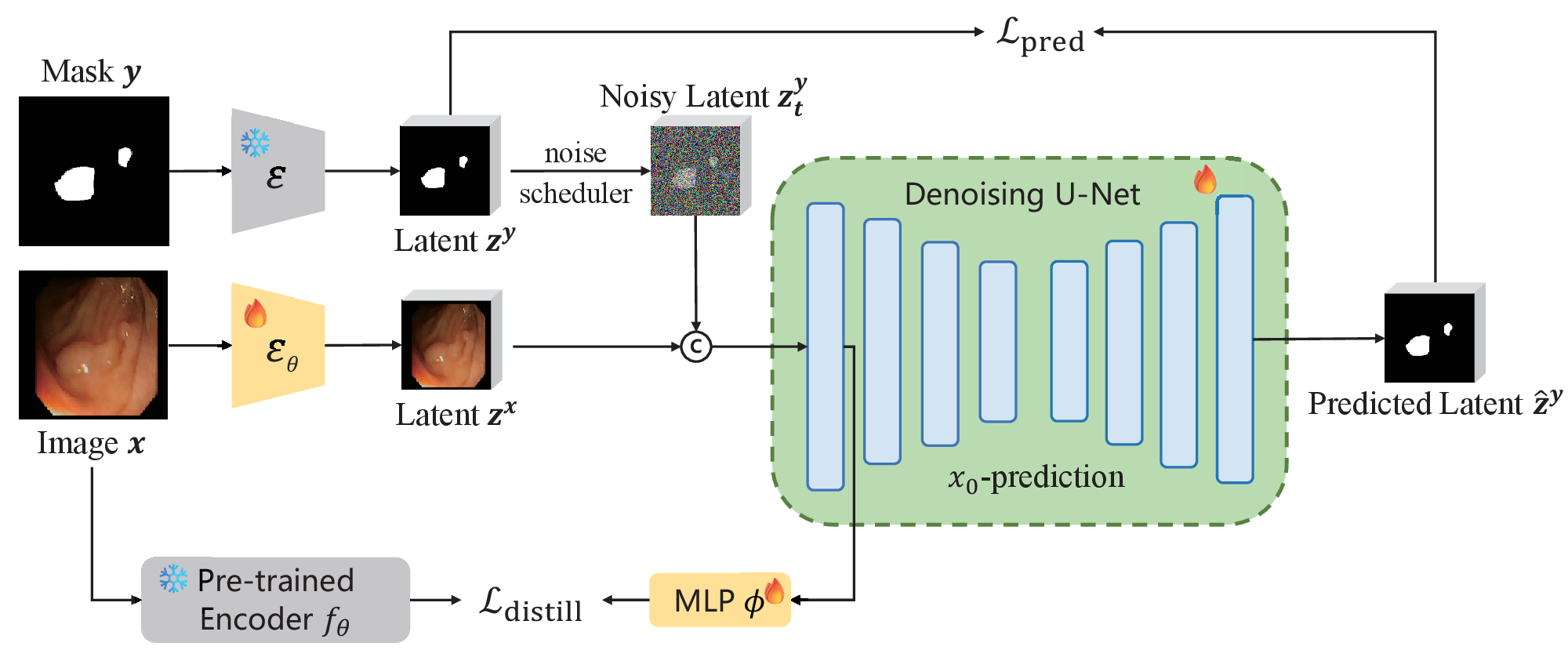}
    \caption{An illustration of training pipeline of LEAF, c means concatenate, the decoder $\mathcal{D}$ is omitted and the noise scheduler is variance-preserving, e.g. $\alpha_{t}^2+\sigma_{t}^2$=1.}
    \label{fig-main-pipeline}
\end{figure}

\section{Methods}

The framework of our method is shown in Figure \ref{fig-main-pipeline}. For the conditioning approach, we follow the process of SDSeg\cite{LinCYYZ24}. Given a Ground-Truth segmentation map $y$, we use a frozen encoder $\mathcal{E}$ to map it into the latent variable $z^{y}$, and add noise via Equation \eqref{eq-add-noise} to obtain the noisy variable $z_{t}^{y}$, where $\epsilon$ is random Gaussian noise, and $\alpha_{t}$ and $\sigma_{t}$ are a set of hyperparameters pre-defined by the noise scheduler; typically, $\alpha_{t}$ decreases while $\sigma_{t}$ increases. For the image $x$, we use a learnable encoder $\mathcal{E}_{\theta}$, initialized with the weights of $\mathcal{E}$, to map it into $z^{x}$. Then, we concatenate $\mathrm{concat}(z^{x};z_{t}^{y})$ as the input to the U-Net and obtain the output $\hat{z}^{y}$.

\begin{equation}
    z_{t}=\alpha_{t}z+\sigma_{t}\epsilon \label{eq-add-noise}
\end{equation}

\subsection{Parameterization Types}

Current diffusion models are used primarily for generation tasks, and prediction objectives generally utilize the following two approaches: (1) $\epsilon$-prediction\cite{NEURIPS2020}, in which the model learns to predict the noise $\epsilon$; (2) $\mathbf{v}$-prediction\cite{SalimansH22}, in which the model learns to predict the velocity defined by Equation \eqref{eq-v-prediction}.

\begin{equation}
    \mathbf{v}:= \alpha_{t}\epsilon-\sigma_{t}z
    \label{eq-v-prediction}
\end{equation}

These parameterization types can be used to estimate the original image via the formula in Equation \eqref{eq-recon}:

\begin{equation}
    \hat{z}=\begin{cases}
        (z_{t}-\sigma_{t}\hat{\epsilon})/\alpha_{t} & ,\textrm{If\ } \epsilon\textrm{-prediction} \\
        \alpha_{t}z_{t}-\sigma_{t}\hat{\mathbf{v}} & ,\textrm{If\ } \mathbf{v}\textrm{-prediction} \\
        \hat{z} & ,\textrm{If\ }x_{0}\textrm{-prediction}
    \end{cases}
    \label{eq-recon}
\end{equation}

Since we freeze the decoder $\mathcal{D}$ during training, the error in $\hat{x}$ originates primarily from $\hat{z}$. Note that both parameterization methods involve a coefficient based on variance when reconstructing $z_{t}$; as $t\rightarrow T$, $\sigma_{t}$ increases while $\alpha_{t}$ decreases. In diffusion models that employ a single-step reverse process at $t=T$, this further amplifies the error in the estimation.

Therefore, we propose that the $x_{0}$-prediction\cite{NEURIPS2020} approach is more suitable for image segmentation. Using $x_{0}$-prediction, the diffusion model directly outputs $\hat{z}$ without introducing additional scaling coefficients, thereby avoiding unnecessary errors. Compared to the other two prediction methods, this approach yields more stable and accurate results. In summary, for a diffusion model pre-trained with either $\epsilon$-prediction or $\mathbf{v}$-prediction, we fine-tune it directly using $x_{0}$-prediction so that it directly predicts the segmentation map. The corresponding loss is shown in the Equation \eqref{eq-loss-pred}, using the $L1$ loss as implemented in SDSeg.

\begin{equation}
    \mathcal{L}_{\textrm{pred}}=\mathcal{L}_{L1}(\hat{z}^{y},z^{y})
    \label{eq-loss-pred}
\end{equation}

\subsection{Features Alignment}

In recent studies, enhancing the ability of diffusion models to extract features is typically achieved by modifying the model architecture. As demonstrated in TransUNet\cite{chen2024transunet} and Diff-Trans\cite{ChowdaryY23}, Transformer architectures can effectively improve the encoder's feature extraction capability, but they also significantly increase the number of model parameters and computational cost. To encourage a U-Net based latent diffusion model to learn rich representations, we introduce a regularization strategy to augment the capacity of convolution, enabling it to capture the representations learned by Transformer architectures.

Inspired by REPA\cite{yu2024repa}, we utilize a pre-trained self-supervised powerful visual encoder $f_{\theta}$, such as DINOv2\cite{OquabDMVSKFHMEA24} or CLIP\cite{RadfordKHRGASAM21}, as the base model for providing robust visual representations. It takes a clean image $x$ as input and produces hidden states $h=f_{\theta}(x)\in\mathbb{R}^{L\times D}$, where $L$ is the number of patches and $D$ is the embedding dimension. In the encoder block of the denoising U-Net, we obtain a feature map $m\in\mathbb{R}^{C\times H\times W}$ and reshape it into $\mathbb{R}^{HW\times C}$, with the condition that $HW=L$. Then, we use a multilayer perceptron (MLP) $\phi$ to project $m$, yielding $\phi(m)\in\mathbb{R}^{L\times D}$, and compute a distillation loss based on cosine similarity:
\begin{equation}
    \mathcal{L}_{\textrm{distill}}=\mathcal{L}(h,\phi(m))=-\frac{1}{N}\sum_{i=1}^{N}(\frac{h_{i}\cdot m_{i}}{\Vert h_{i}\Vert\cdot \Vert m_{i}\Vert})
    \label{eq-loss-distill}
\end{equation}

We add this loss to the prediction loss above, and the total objective loss is shown in Equation\eqref{eq-loss-distill}, where $\lambda$ is a positive constant controlling the strength of the distillation alignment.
\begin{equation}
\mathcal{L}_{\textrm{total}}=\mathcal{L}_{\textrm{pred}}+\lambda\mathcal{L}_{\textrm{distill}}
    \label{eq-loss-total}
\end{equation}

\subsection{Inference}

During inference, we initialize the segmentation map with standard Gaussian noise $z_{T}^{y}$, and encode the input image into $z^{x}$ using $\mathcal{E}_{\theta}$. The concatenated features $(z_{T}^{y};z^{x})$ are then fed into the U-Net. Notably, we remove the pre-trained visual encoder and MLP during this phase, ensuring no additional parameters are introduced compared to the original model. Following SDSeg\cite{LinCYYZ24}, we apply a single-step reverse process to obtain $\hat{z}^{y}$, which is then decoded to the pixel space via $\mathcal{D}$ to produce the final segmentation map.

\section{Experimental Results}

\subsection{Experimental setup}

\subsubsection{Datasets and Evaludation Metrics}
To comprehensively evaluate the proposed method, we conduct experiments on four public medical image segmentation tasks: (1) Optic-cup segmentation from retinal fundus images (REFUGE2 (REF)\cite{OrlandoFBKBDFHK20}), (2) Polyp segmentation from colonoscopy images (CVC-ClinicDB (CVC)\cite{BernalSFGRV15}), (3) COVID-19 lesion segmentation (QaTa-Covid19 (Qata)\cite{DegerliKCG22}), and (4) Skin lesion segmentation from dermoscopy images (ISIC 2018\cite{CodellaGCHMDKLM18,Tschandl2018DescriptorT}). We use mean Dice and mean IoU as primary evaluation metrics. For REFUGE2, we used the data partition defined in SDSeg. For ISIC 2018, we adopted a training-testing ratio of $7:3$, while for CVC, we utilized an 80:10:10 data partition. QaTa, on the other hand, used the default training and testing sets.


\subsubsection{Implementation Details} We implemented LEAF using the PyTorch platform and trained/evaluated on a single NVIDIA A800 GPU.
All images were resized to a resolution of $256\times 256$. The pre-trained unconditional latent diffusion model was based on LDM-KL-8\cite{RombachBLEO22}. To optimize the model, we utilized the standard AdamW optimizer with a batch size of $4$. The learning rate was set to $4\times 10^{-5}$ with a warm-up constant learning rate scheduler. To handle the concanated input, we duplicated the U-Net input layer from 4 channels to 8 channels and initialized its weights by halving the original weights as mentioned in Marigold\cite{KeOHMDS24}. The pre-trained vision encoder was DINOv2\cite{OquabDMVSKFHMEA24}.

\subsection{Performance Comparison}

We conducted extensive experiments in various evaluated datasets to assess the effectiveness of LEAF, as shown in Table \ref{tab-performance}. LEAF represented a generic approach for latent diffusion models without domain-specific modules tailored to particular medical imaging modalities. Consequently, our comparisons focused on models with strong generalization capabilities. For a fair comparison with SDSeg, we re-trained it under our framework using the same configurations. MedSegDiff was re-evaluated on CVC. The results of QaTa and ISIC2018 were directly cited from MMDSN\cite{10822810} and BGDiffSeg\cite{Guo_BGDiffSeg_MICCAI2024}, and REF from SDSeg\cite{LinCYYZ24}, CVC from Diff-Trans\cite{ChowdaryY23}.

\begin{table}[htb!]
    \centering
    \caption{Performance comparison on our proposed model and existing SOTA medical segmentation models.}
    \label{tab-performance}
    \begin{tabular}{c|c|cc|cc|cc|cc}
    \hline
    & \multirow{2}{*}{Model} & \multicolumn{2}{c|}{REF} & \multicolumn{2}{c|}{CVC} & \multicolumn{2}{c|}{QaTa} & \multicolumn{2}{c}{ISIC2018}\\
    \cline{3-10} & & Dice & IoU & Dice & IoU & Dice & IoU & Dice & IoU\\
    \hline
    \multirow{3}{*}{\makecell{CNN/Transformer-based}} & U-Net\cite{RonnebergerFB15} & 80.1 & - & 85.6 & 80.5 & 79.0 & 69.5 & 87.6 & 77.9 \\
    & TransUNet\cite{chen2024transunet} & 85.6 & - & 92.0 & 87.8 & 78.6 & 69.1 & 88.7 & 79.7 \\
    & Swin-UNet\cite{CaoWCJZTW22} & 84.3 & - & 91.4 & 87.4 & 78.1 & 68.3 & - & - \\
    \hline
    \multirow{2}{*}{Diffusion-based} & MedSegDiff\cite{MedSegDiff} & 86.3 & 78.2 & 92.4 & 88.9 & 76.5 & 67.2 & 85.5 & 74.7 \\
        & SDSeg\cite{LinCYYZ24} & 88.7 & 80.9 & 93.6 & 89.3 & 77.6 & 68.0 & 88.1 & 79.7 \\
    \hline
    Proposed & LEAF & \textbf{89.5} & \textbf{81.5} & \textbf{95.2} & \textbf{90.9} & \textbf{80.2} & \textbf{71.0} & \textbf{90.5} & \textbf{84.1} \\
    \hline
    \end{tabular}
\end{table}

As shown in Table \ref{tab-performance}, LEAF outperforms all baseline models on datasets involving various types of medical image, validating its effectiveness and generalizability. While sharing the same U-Net architecture as SDSeg, our method replaces its parameterization type and aligns convolutional layers with features extracted from a Transformer-based encoder, achieving significant performance improvements.

\subsection{Ablation Study}

\subsubsection{Ablation for fine-tuning pipeline} We establish the baseline as the model using the original $\epsilon$-prediction without feature alignment (first row in Table \ref{tab-ablation}). From the table, we observe that merely changing the prediction method from $\epsilon$-prediction to $\mathbf{v}$-prediction yields significant performance gains. Furthermore, switching to $x_{0}$-preditcion without scaling factors further improves model performance. Finally, feature alignment achieves the highest performance compared to other configurations. Although these features come from different model architectures and DINOv2 is not fine-tuned on medical images, it can still improve the segmentation performance. We emphasize that these improvements are consistent across all evaluated datasets, with nontrivial magnitude.

\begin{table}[htb!]
    \centering
    \caption{Ablation study for parameterization and features alignment. The rows with gray color highlight the features of the model distilled from the vision encoder during training for clearer comparison.}
    \label{tab-ablation}
    \begin{tabular}{c|c|cc|cc|cc|cc}
    \hline
    \multirow{2}{*}{\makecell{Parameterization \\ Types}} & \multirow{2}{*}{\makecell{Feature\\ Alignment}} & \multicolumn{2}{c|}{REF} & \multicolumn{2}{c|}{CVC} & \multicolumn{2}{c|}{QaTa} & \multicolumn{2}{c}{ISIC2018}\\
    \cline{3-10} & & Dice & IoU & Dice & IoU & Dice & IoU & Dice & IoU\\
    \hline
    $\epsilon$-prediction & \ding{55} & 88.47 & 79.59 & 90.15 & 83.68 & 74.27 & 63.80 & 87.67 & 80.13 \\
    \rowcolor{gray!20}$\epsilon$-prediction & \ding{51} & 87.61 & 78.43 & 91.63 & 87.10 & 74.56 & 63.97 & 87.34 & 79.48 \\
    $\mathbf{v}$-prediction & \ding{55} & 89.21 & 80.92 & 93.75 & 89.32 & 79.10 & 69.74 & 90.35 & 83.91 \\
    \rowcolor{gray!20}$\mathbf{v}$-prediction & \ding{51} & 89.30 & 80.88 & 94.89 & 90.44 & 79.32 & 69.98 & 90.52 & 84.11 \\
    $x_{0}$-prediction & \ding{55} & 89.21 & 79.08 & 94.49 & 89.94 & 79.08 & 69.85 & 90.39 & 83.94 \\
    \rowcolor{gray!20}$x_{0}$-prediction & \ding{51} & 89.53 & 81.45 & 95.17 & 90.94 & 80.15 & 71.04 & 90.54 & 84.18 \\
    \hline
    \end{tabular}
\end{table}

\subsubsection{Ablation for Feature Alignment}
We investigate the hyperparameter $\lambda$ that controls alignment strength and the model size of the vision encoder, with results shown in Table \ref{tab-lambda}. Firstly, the optimal value of $\lambda$ generally varies across different datasets, which may be related to the distribution and inherent characteristics of the data. Secondly, the impact of different $\lambda$ values on model performance is not significant, for example, the maximum absolute difference of dice score on ISIC2018 is $0.3$, while on QaTa, it is over $1.0$. We believe this is due to the fact that the images in QaTa  contain more structural information, making them more sensitive to the distillation strength. Overall, the model performs better with $\lambda>0$ than it does with $\lambda=0$, further validating the effectiveness of feature alignment.

\begin{table}[htb!]
    \centering
    \caption{The effect of hyperparameter $\lambda$ for features alignment.}
    \label{tab-lambda}
    \begin{tabular}{c|cc|cc|cc|cc}
    \hline
    \multirow{2}{*}{$\qquad\lambda\qquad$} & \multicolumn{2}{c|}{REF} & \multicolumn{2}{c|}{CVC} & \multicolumn{2}{c|}{QaTa} & \multicolumn{2}{c}{ISIC2018}\\
    \cline{2-9} & Dice & IoU & Dice & IoU & Dice & IoU & Dice & IoU\\
    \hline
    $0$ & 89.21 & 79.08 & 94.49 & 89.94 & 79.08 & 69.85 & 90.39 & 83.94 \\
    $0.15$ & 89.39 & 81.19 & 95.07 & 90.77 & 79.62 & 70.37 & 90.24 & 83.83 \\
    $0.25$ & 89.41 & 81.24 & 94.97 & 90.57 & 79.74 & 70.65 & \textbf{90.54} & \textbf{84.06} \\
    $0.50$ & 89.33 & 81.12 & 94.21 & 89.35 & 80.05 & 70.87 & 90.43 & 84.06 \\
    $0.75$ & \textbf{89.53} & \textbf{81.45} & 95.01 & 90.67 & 79.98 & 70.93 & 90.51 & 84.05 \\
    $1.0$ & 89.44 & 81.27 & \textbf{95.17} & \textbf{90.94} & 79.87 & 70.77 & 90.34 & 83.90 \\
    $1.25$ & 89.43 & 81.27 & 95.07 & 90.76 & \textbf{80.15} & \textbf{71.04} & 90.30 & 83.93 \\
    \hline
    \end{tabular}
\end{table}

\subsection{Quality Results}

\subsubsection{Stability} Diffusion models are non-deterministic models; thus, many previous models have attempted to reduce segmentation uncertainty by running the model multiple times and ensembling the results in some way as the final outcome, which increases the inference speed of the model. The segmentation results for LEAF are obtained from a single run, so it is necessary to demonstrate their stability, with the results presented in Table \ref{tab-stability}.

\begin{table}[htb!]
    \centering
    \caption{Stability experiments. We selected $10$ different random seeds to inference $10$ times and calculated the standard deviation of the Dice score (\%).}
    \label{tab-stability}
    \begin{tabular}{c|c|c|c|c}
    \hline
    Parameterization Types & {\quad REF\quad} & {\quad CVC\quad} & {\quad QaTa\quad} & {\quad ISIC2018\quad} \\
    \hline
    $\epsilon$-prediction & 0.09 & 0.32 & 0.14 & 0.29 \\
    $x_{0}$-prediction & 0.05 & 0.11 & 0.08 & 0.06 \\
    \hline
    \end{tabular}
\end{table}

The experimental results demonstrated that the $\epsilon$-prediction method indeed has a larger variance compared to the $x_{0}$-prediction method. Moreover, the standard deviation of the models trained using the $x_{0}$-prediction method is mostly on the order of $10^{-2}$, indicating a very small difference. This difference is significantly smaller than the improvement brought about by our proposed method, further proving the effectiveness and stability of our approach.

\subsubsection{Visualization} 
\begin{figure}[htb]
    \centering
    \includegraphics[scale=0.16]{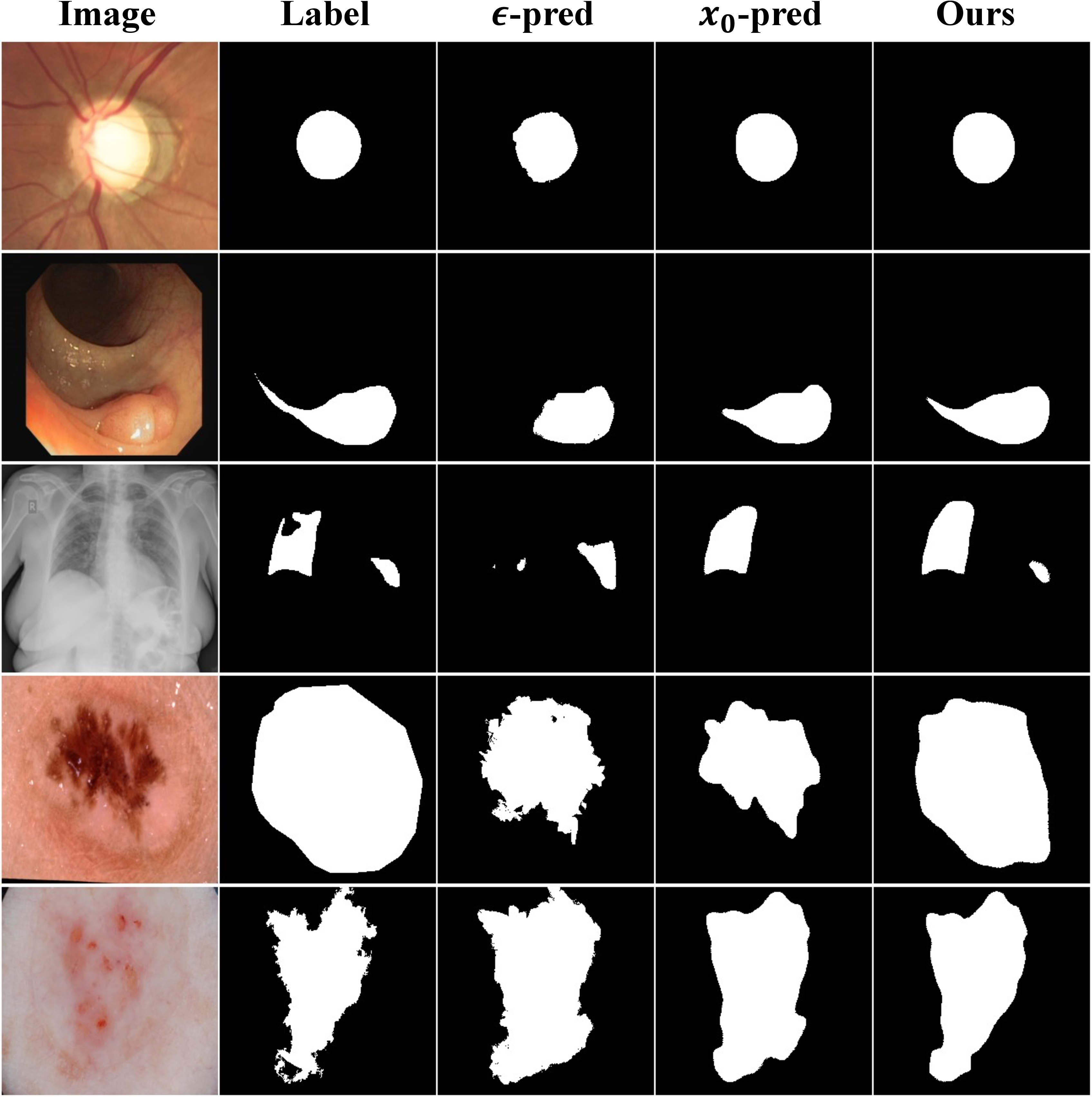}
    \caption{Visualization of segmentation results.}
    \label{fig-vis}
\end{figure}

Additionally, we visualize the segmentation results on different medical segmentation tasks, as shown in Figure \ref{fig-vis}.

\section{Conclusion}

In this paper, we propose LEAF, an efficient and generalized framework for fine-tuning a latent diffusion model for medical image segmentation. We investigate the effect of parameterization type and propose to use $x_{0}$-prediction parameterization for segmentation task. We also introduce a simple featrue alignment method via distilling vision encoder, providing a better representation for the CNN-based U-Net. LEAF brings about zero cost for inference and can be easily adapted to other LDM-based segmentation models.

\begin{credits}
\subsubsection{\ackname} This study was funded by the Guangdong S\&T Program (Grant No. 2024B0101040005),  the National Natural Science Foundation of China (Grant No. 62461146204), the National Key Research and Development Program of China (Grant No. 2021YFB0300103), and the Pazhou Lab program (Grant No. PZL2023KF0001.)

\subsubsection{\discintname}
The authors have no competing interests to declare
that are relevant to the content of this article.
\end{credits}

%
%
%
\bibliographystyle{splncs04}
\bibliography{Paper-2956}

\newpage
\appendix

\section{Stability of Feature Alignment}

We evaluate the stability of the hyperparameter $\lambda$ by testing different random seeds on the QaTa dataset. The results indicate that minor performance fluctuations are primarily due to randomness, confirming the robustness of $\lambda$.

\begin{table}[htb!]
    \centering
    \caption{Dice scores under different values of $\lambda$}
    \label{tab-lambda-stability}
    \begin{tabular}{c|c|c|c|c|c|c}
    \hline
    $\lambda$ & $0.15$ & $0.25$ & $0.5$ & $0.75$ & $1.0$ & $1.25$ \\
    \hline
    Dice & $79.88\pm 0.11$ & $79.82\pm 0.14$ & $79.81\pm 0.06$ & $79.89\pm 0.01$ & $80.06\pm 0.01$ & $80.17\pm 0.01$ \\
    \hline
    \end{tabular}
\end{table}

\section{Effect of Noise Scheduler}

We also study the impact of different noise schedulers on segmentation performance. The results suggest that scheduler choice significantly influences model outcomes, warranting further investigation to determine the optimal configuration.

\begin{table}[h]
\centering
\caption{Comparison of beta schedules and corresponding Dice scores}
\label{tab:beta_schedule}
\begin{tabular}{c|c|c|c|c|c}
\hline
\multirow{2}{*}{Beta Schedule} & linear & linear & linear & scaled linear & scaled linear \\
\cline{2-6} & 0.0001--0.02 & 0.00085--0.012 & 0.0015--0.0155 & 0.0001--0.02 & 0.0015--0.0155 \\
\hline
Dice & $79.58$ & $76.29$ & $79.59$ & $77.64$ & $77.70$ \\
\hline
\end{tabular}
\end{table}

\end{document}